\pdfoutput=1
\documentclass[letterpaper, 10 pt, conference]{ieeeconf}
\IEEEoverridecommandlockouts                              

\usepackage[utf8]{inputenc}

\usepackage{graphics} 
\usepackage{times} 
\usepackage{amsmath} 
\usepackage{amssymb}  

\usepackage{cite}
\usepackage[dvipsnames]{xcolor}

\usepackage{algorithm}
\usepackage[noend]{algpseudocode}

\usepackage{color}
\usepackage{tikz}
\usepackage{comment}
\usepackage{hyperref}

\title{Onboard View Planning of a Flying Camera for High Fidelity 3D Reconstruction of a Moving Actor}
\author{Qingyuan Jiang$^{1}$, and Volkan Isler$^{1}$
    \thanks{*This work is supported by the NSF NRI Grant \#2022894.}
    \thanks{$^{1}$Qingyuan Jiang and Volkan Isler are with Department of         Computer Science, University of Minnesota, Twin Cities, Minneapolis, MN, 55455
    {\tt\small \{jian0345, isler\}@umn.edu}}
}

\date{March 2023}

\begin{document}

\maketitle

\begin{abstract}

Capturing and reconstructing a human actor's motion is important for filmmaking and gaming. Currently, motion capture systems with static cameras are used for pixel-level high-fidelity reconstructions. Such setups are costly, require installation and calibration and, more importantly, confine the user to a predetermined area. 
In this work, we present a drone-based motion capture system that can alleviate these limitations. We present a complete system implementation and study view planning which is critical for achieving high-quality reconstructions. The main challenge for view planning for a drone-based capture system is that it needs to be performed during motion capture. To address this challenge, we introduce simple geometric primitives and show that they can be used for view planning.
Specifically, we introduce Pixel-Per-Area (PPA) as a reconstruction quality proxy and plan views by maximizing the PPA of the faces of a simple geometric shape representing the actor.
Through experiments in simulation, we show that PPA is highly correlated with reconstruction quality. We also conduct real-world experiments showing that our system can produce dynamic 3D reconstructions of good quality. We share our code for the simulation experiments in the link: \href{https://github.com/Qingyuan-Jiang/view_planning_3dhuman}{\text{https://github.com/Qingyuan-Jiang/view\_planning\_3dhuman}}.
\end{abstract}

\section{INTRODUCTION}  \label{sec:introduction}
Capturing a human actor's motion with fine details is essential due to its applications in virtual reality and related metaverse applications. However, obtaining such \emph{high-fidelity} reconstructions remains a challenging problem~\cite{fuchs_virtual_1994, lanier_virtually_2001} due to the actor's motion. Recently developed multi-camera systems can generate reconstructions at the sub-pixel level of hand details or facial gestures~\cite{joo_panoptic_2015, yu_humbi_2020}. The primary limitation of these systems is that they rely on stationary, pre-calibrated cameras which confine the user to a motion capture area.
In this work, we tackle the challenge of acquiring images with drones for the high-fidelity reconstruction of a dynamic human actor. We focus on the view planning strategy to improve the reconstruction quality.

There are many technical challenges in designing such a view planning algorithm when the actor is dynamic: 
At the algorithmic level, the planning space is heavily enlarged due to the uncertainty of the actor's movement. Also, since the target surface keeps changing, there is no prior information and no closed-form objective functions for the planning algorithm.
At the system level, it requires a real-time capability to plan views for a dynamic object. However, for a standard drone system, as we have in this work DJI M100 with Nvidia Jetson TX1 with 4G memory, it is costly to acquire, store, and communicate RGB-D images even in $640\times480$ resolution and $15$Hz, which brings delays to the system and occupies bandwidth and computation resources. The planning decision needs to be made in a short time.

\begin{figure}[th]
    \centering
    \includegraphics[width=0.9\columnwidth]{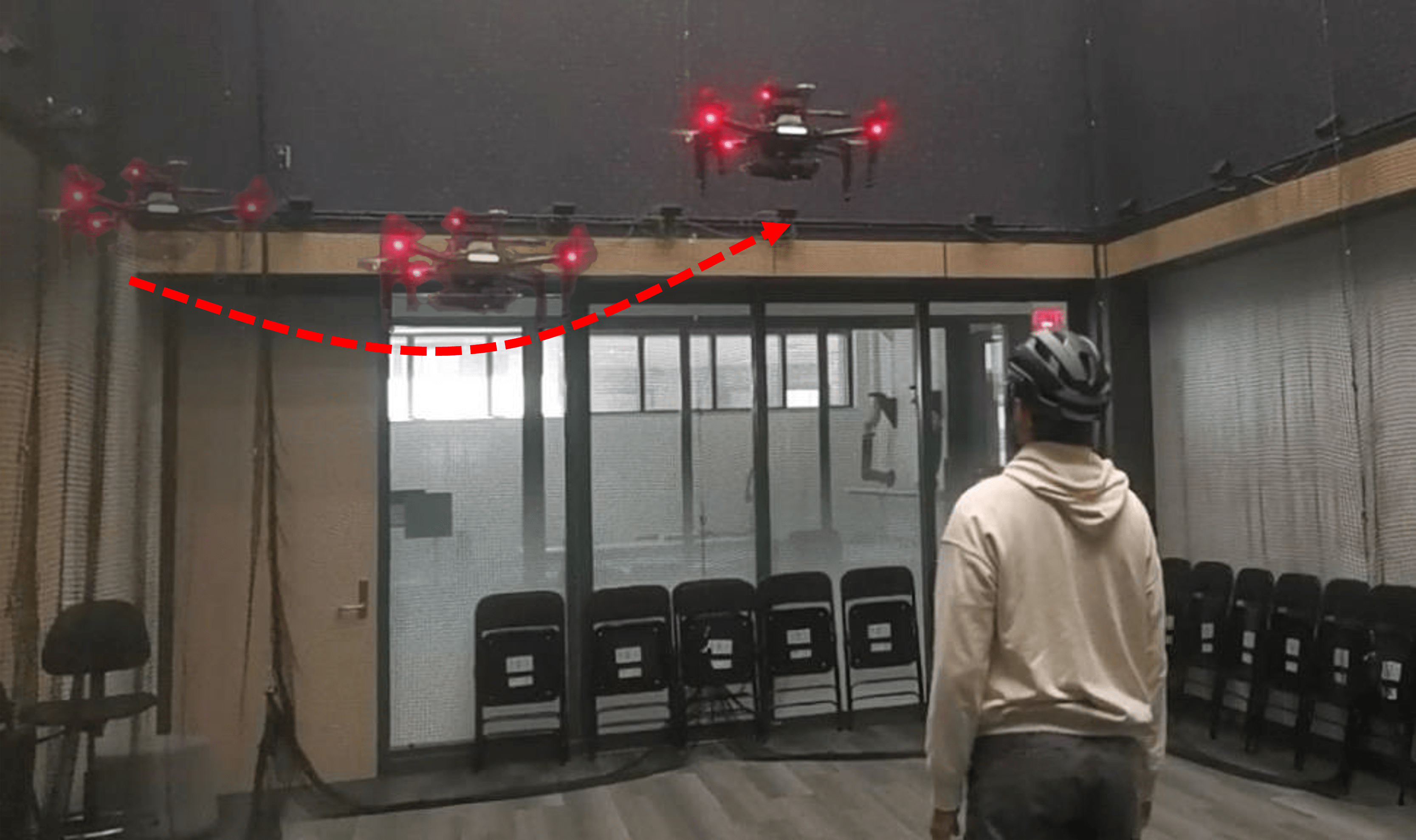}
    \caption{\textbf{Capturing images with a flying camera for the high-fidelity reconstruction of a dynamic actor.} We build a drone system to capture a dynamic actor's high-fidelity 3D reconstruction and study the view planning strategy for better reconstruction quality.}
    \label{fig:teaser}
\end{figure}

We present a view planning strategy that addresses these challenges. We present a new objective function, Pixels-Per-Area (PPA), to measure the fidelity of 3D reconstruction.
To overcome real-time requirements, we propose a view planning algorithm based on our PPA proxy that is computationally moderate  and  uses simple geometric primitives (e.g., cuboids) as intermediate actor models for planning. The view planning objective then becomes that of maximizing the PPA of the faces of the cuboid over time.

Our contributions can be summarized as follows.
\begin{itemize}
    \item
    We propose to use the Pixels-Per-Area (PPA) function as a proxy for reconstruction quality. We formulate the view planning algorithm as the optimization problem of maximizing the PPA and provide a strategy to solve it based on the Jacobian vector.
    \item 
    To validate our algorithm, we build a drone system that can produce high-fidelity 3D reconstructions of a dynamic, moving human actor. To accomplish this task, the system can actively track the dynamic actor, planning and capturing views during flight and reconstructing the actor in high fidelity from camera readings after the flight.
    \item
    We evaluate our view planning algorithm both in simulation and real-world experiments. Our quantitative and qualitative results show that our system can provide a dynamic human 3D reconstruction of good quality in both coverage ratio and Chamfer distance.
\end{itemize}

\section{RELATED WORK} \label{sec:related_work}
Reconstruction with flying cameras has received significant attention. Many works aim for a dynamic human target and a 3D skeleton pose reconstruction. Meanwhile, other works build high-fidelity reconstructions for large-scale static objects, such as buildings.

\subsection{Human pose reconstruction with drones}

There is an increasing amount of work on controlling drones to produce human poses actively. 
For example, the drone is controlled \cite{zhou_human_2018} to orbit around the actor, showing that such movements benefit the actor's skeleton reconstruction.
Viewpoints are either selected from a dome \cite{pirinen_domes_2019}, optimized over the uncertainty of the 3D human pose \cite{kiciroglu_activemocap_2020} to obtain better 3D human pose reconstructions or by optimizing the artistic meaning \cite{bonatti_autonomous_2020, bonatti_batteries_2021}.

On the other hand, some studies extend this setting to multiple drones.
Some of them \cite{alcantara_optimal_2021} focus on multi-UAV trajectory optimization and coordination by assuming the viewing directions are given.
Planning actively for view selection has been studied in \cite{tallamraju_aircaprl_2020, tallamraju_active_2019, saini_markerless_2019, ahmad_aircap_2019}. Tallamraju et al. \cite{tallamraju_active_2019} and Saini et al.\cite{saini_markerless_2019} use MPC to estimate human odometry and then optimize for the 3D body pose.
In recent work, Ho et al. \cite{ho_3d_2021} obtained skeleton reconstructions with formation planning while avoiding obstacles.

Compared to these works, 
our system is designed for high-fidelity reconstructions by using geometric primitives as intermediate actor models and proposing a proxy for reconstruction quality.

\subsection{View planning for high-fidelity 3D reconstruction of static objects}

Researchers also use drones to reconstruct objects on a large scale and in high fidelity. 
Color images from a drone's flight are used to reconstruct a single building in \cite{wefelscheid_three-dimensional_2012, daftry_building_2015}.
The problem is later formulated as a SLAM problem in \cite{li_uav-based_2017} from Li et al. 
Semantic information is added to plan the UAV path in \cite{koch_automatic_2019}. 
Views are also planned in \cite{peng_adaptive_2019, peng_view_2018, peng_visual_2020} by formulating the problem as the Traveling Salesperson Problem with Neighborhoods. 
Later, researchers pushed for a larger area, such as a landslide in \cite{gupta_application_2018}, and a large urban area in \cite{zhang_continuous_2021, wahbeh_automatic_2022}. 

While such UAV systems can build 3D reconstruction results for static targets, producing a dynamic one is intractable because only limited views are available at each time stamp. We show how our system differs from the above ones in Sec.~\ref{sec:system}.


\section{PROBLEM FORMULATION}  \label{sec:formulation}
In this section, we start by representing a human surface and its pose. Later we introduce the model of the drone's pose representation and formulate the problem as an optimization problem with constraints.

\begin{figure}[th]
    \centering
    \includegraphics[width=0.7\columnwidth]{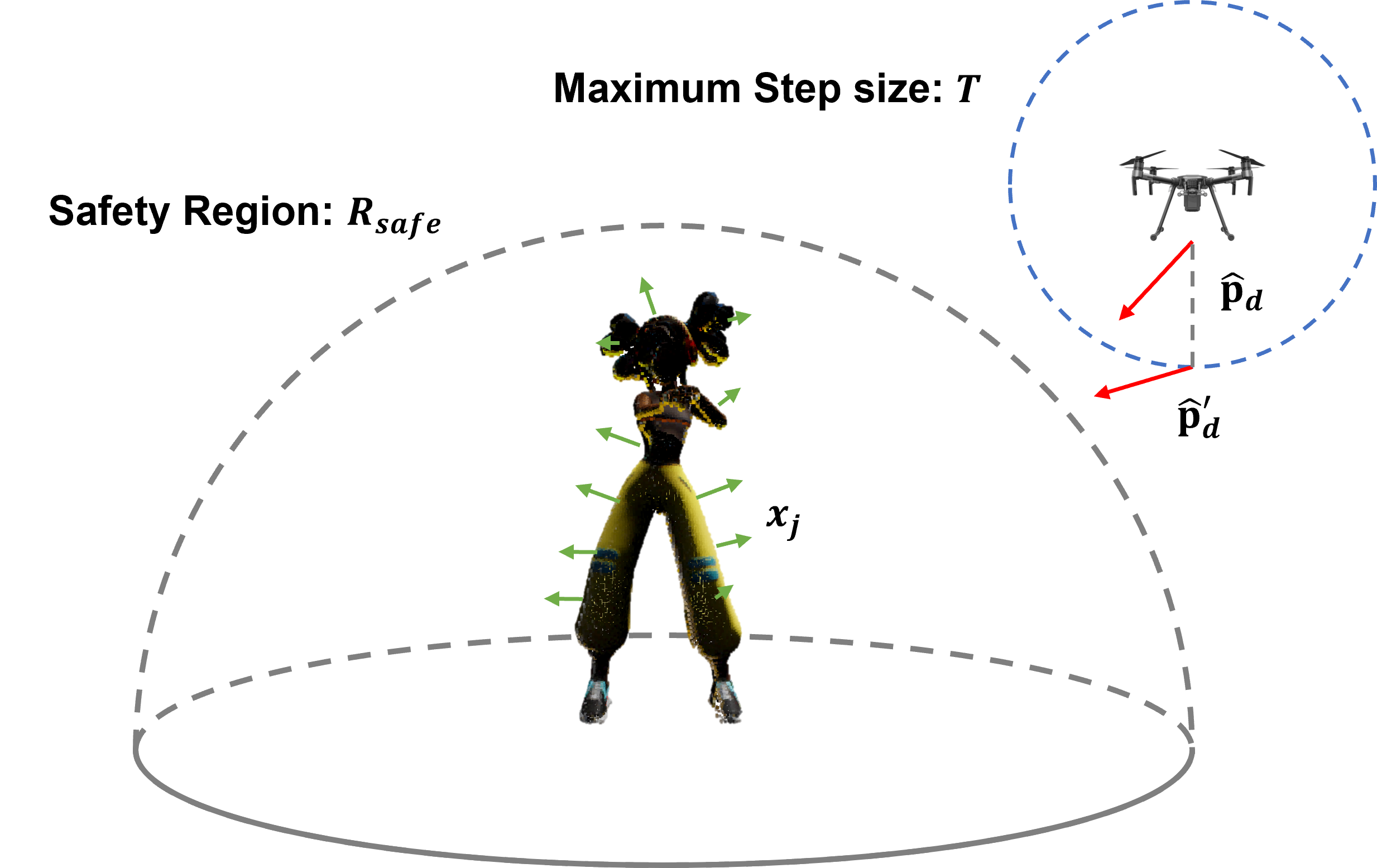}
    \caption{\textbf{Formulation}. We model the actor as a set of geometric primitives with surface normals. We would like to find a viewpoint that minimizes the reconstruction error while maintaining a minimum distance to ensure the actor's safety.}
    \label{fig:formulation}
\end{figure}

\subsection{Human representation}  \label{sec:PPA_def}

We model the actor as a set of patches. A patch is defined as a planar surface of bounded size, represented with i)~a point at its centroid and ii)~a normal vector indicating the orientation. For example, a triangle from a mesh can be treated as a patch in the highest resolution.

Suppose we have $\mathbf{m}$ patches from the actor to visit in a 3-Dimensional space. Mathematically, we use the geometric center as its representation point. We use $\mathbf{p}_j \in \mathbb{R}^3$ to denote the position of the point with index $j$, and use $\mathbf{n}_j \in \mathbb{R}^3$ to denote the normal vector of the patch. The pose of the $j$-th patch is denoted by $\mathbf{x}_j \in \mathbb{R}^6$, where $\mathbf{x}_j = \left( \mathbf{p}_j, \mathbf{n}_j \right)$. The actor is modeled as the set of patches $\mathcal{X} = \{ \mathbf{x}_j \}$. Meanwhile, We use $\mathbf{x}_a = \left( \mathbf{p}_a, \mathbf{n}_a \right)$ to denote the actor's pose as a whole. Note that all normal vectors has normalized length, i.e. $\|\mathbf{n}_j\|=1$, $\|\mathbf{n}_a\|=1$. By default, we use the L2 norm with the notation $\|\cdot\|$.
Also, because we normalize all patch's area when the usage in Sec.~\ref{sec:PPA_def}, we do not define the area for each patch here.

\subsection{Drone's pose representation}
The localization of a drone's pose can be highly inaccurate in many circumstances. We use $\mathbf{x}_d = \left( \mathbf{p}_d, \mathbf{n}_d \right)$ to denote the ground truth of the drone's pose. We denote its estimation as $\mathbf{\hat{x}}_d = \left( \mathbf{\hat{p}}_d, \mathbf{\hat{n}}_d \right)$.
From the current drone's pose $\mathbf{\hat{x}}_d$, we estimate the actor as a set of patches. We denote it as $\mathcal{X}_{est} = \{\mathbf{\hat{x}}_j\}$. Similarly, we have $\|\mathbf{n}_d\|=\|\mathbf{\hat{n}}_d\|=\|\mathbf{\hat{n}}_j\| = 1$.

\subsection{Formulation}
Given an estimation of the drone's pose $\mathbf{\hat{x}}_d$ and actor observation $\mathcal{X}_{est}$, we would like to compute a new view point $ \mathbf{\hat{x}}_d^{'} = \left( \mathbf{\hat{p}}_d^{'}, \mathbf{\hat{n}}_d^{'} \right) $  in a local area  with better reconstruction quality. Before that, we need to define the safety region and the maximum step size.

\subsubsection{Safety regions} \label{sec:safety_region} To ensure the actor's safety, we define a hemispherical space around the actor as the safe region. We use $R_{\text{safe}} > 0$ to denote the radius of the hemispherical space. The distance between the updated viewpoint and the actor's position should always be greater than $R_{\text{safe}}$.

\subsubsection{Maximum step size} \label{sec:step_size} To constraint the new viewpoint close to the current estimation, we define a maximum step size $T$, such that $\| \mathbf{\hat{p}}_d^{'} -  \mathbf{\hat{p}}_d \| \leq T$.

\subsubsection{Formulation} \label{sec:formulation_equation}
Now that we are ready to formulate the problem. Given an estimation of camera pose $\mathbf{\hat{x}}_d$ and an estimation of the actor model from reconstruction $\mathcal{X}_{est}$, we would like to find a new viewpoint $\mathbf{\hat{x}}_d^{'}$ within step size $T$ and out of safety region, such that the reconstruction error is minimized.

\begin{equation} \label{eq:obj_func}
    \begin{split}
    \mathbf{\hat{x}}_d^{'} = \arg \min_{\mathbf{\hat{x}}_d^{'}} \quad & \sum_j \| \mathbf{x}_j -  \mathbf{x}_j^{'} \| \\
    \text{s.t. }    \quad &   \| \mathbf{\hat{p}}_d^{'} - \mathbf{\hat{p}}_d \| \leq T \\
    &   \|\mathbf{\hat{p}}_d^{'} - \mathbf{p}_a\| \geq R_{\text{safe}}
    \end{split}
\end{equation}

We will solve the formulated problem above with our view planning module described in Sec.~\ref{sec:view_planning}.

\section{VIEW PLANNING METHODOLOGY} \label{sec:approach} \label{sec:view_planning}
Because we do not have the ground truth of human patches $\mathbf{x}_j$ in the formulation Eq.~\ref{eq:obj_func}, we define Pixels-Per-Area (PPA) in this part as a proxy for the reconstruction quality.

\subsection{Pixels-Per-Area (PPA) as reconstruction quality proxy}

We define the Pixels-Per-Area (PPA) as the projection area in the image plane of a 3D patch, which is a function of the drone's pose $\mathbf{x}_d$ and the pose of a patch $\mathbf{x}_j$:

\begin{equation} \label{eq:ppa_define}
    \mathbf{ppa} \left( \mathbf{x}_d, \mathbf{x}_j \right) 
    = \frac{\cos(\alpha(\mathbf{n}_d, \mathbf{n}_j))}{d(\mathbf{p}_d, \mathbf{p}_j)}
\end{equation}

$\alpha(\mathbf{n}_d, \mathbf{n}_j)$ defines the acute angle between $\mathbf{n}_j$ and $\mathbf{n}_d$. $d(\mathbf{p}_d, \mathbf{p}_j)$ defines the Euclidean distance between the drone and the patch given as: 

\begin{equation} \label{eq:ppa_note}
    \begin{split}
        \cos\left(\alpha(\mathbf{n}_d, \mathbf{n}_j)\right) & = \frac{\mathbf{n}_d \cdot \mathbf{n}_j}{\| \mathbf{n}_d \| \cdot \|\mathbf{n}_j\|}     \\
        d(\mathbf{p}_d, \mathbf{p}_j) & = \| \mathbf{p }_d - \mathbf{p}_j \|
    \end{split}
\end{equation}

\begin{figure}[ht]
    \centering
    \includegraphics[width=0.6\columnwidth]{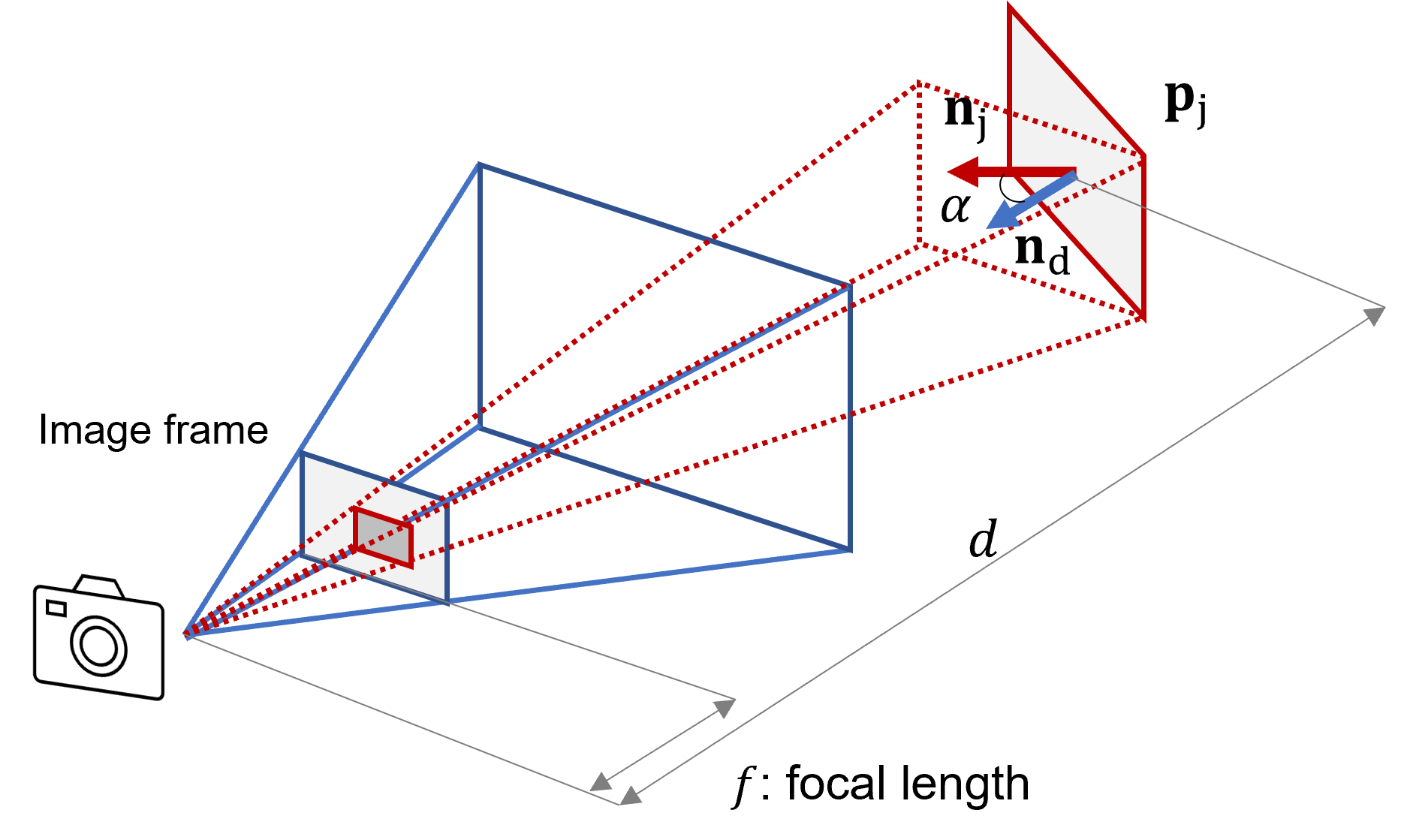}
    \caption{\textbf{Geometric meaning of the PPA value}. We define the PPA value of a patch as the ratio between the projection area in the image plane (colored in \textcolor{blue}{blue}) and the patch's original area. (colored in \textcolor{red}{red}). $f$: the focal length of a camera.}
    \label{fig:ppa_geometry_meaning}
\end{figure}

As its name implies, the PPA function describes the pixels occupied in the image plane by a 3D area.
We include a detailed explanation of the PPA's geometry meaning in our supplementary material.
With the PPA function, we re-write the objective function as below in Eq.~\ref{eq:obj_func_ppa}.

\begin{equation} \label{eq:obj_func_ppa}
    \begin{split}
    \mathbf{\hat{x}}_d^{'} = \max_{\mathbf{\hat{x}}_d^{'}} \quad & \sum_j \mathbf{ppa} \left( \mathbf{\hat{x}}_d^{'}, \mathbf{\hat{x}}_j \right)\\
    \text{s.t. }    \quad &   \| \mathbf{\hat{p}}_d^{'} - \mathbf{\hat{p}}_d \| \leq T\text{,} \quad \|\mathbf{\hat{p}}_d^{'} - \mathbf{p}_a\| \geq R_{\text{safe}}
    \end{split}
\end{equation}

\subsection{View planning based on PPA}

We make a one-step update on the drone's pose estimation by maximizing the summation of PPA values throughout patches with Eq.~\ref{eq:obj_func_ppa}. To do so, we calculate the gradient from the Jacobian vector with respect to the drone's pose estimation as described in Eq.~\ref{eq:ppa_jacob_pose}, whose components are given in Eq.~\ref{eq:ppa_jacob_pos} and Eq.~\ref{eq:ppa_jacob_ori}.

\def\verticaldistance{15pt}
\begin{equation}\label{eq:ppa_jacob_pose}
    \frac{\partial}{\partial \mathbf{\hat{x}}_d} \sum_j \mathbf{ppa} \left( \mathbf{\hat{x}}_d^{'}, \mathbf{\hat{x}}_j \right)  \\
    = \sum_j
    \begin{bmatrix}
        \dfrac{\partial}{\partial \mathbf{\hat{p}}_d} 
            \mathbf{ppa} \left( \mathbf{\hat{x}}_d^{'}, \mathbf{\hat{x}}_j \right) \\[\verticaldistance]
        \dfrac{\partial}{\partial \mathbf{\hat{n}}_d} 
            \mathbf{ppa} \left( \mathbf{\hat{x}}_d^{'}, \mathbf{\hat{x}}_j \right) \\
    \end{bmatrix}
\end{equation}



\begin{align}
    & \frac{\partial}{\partial \mathbf{\hat{p}}_d} \sum_j \mathbf{ppa} \left( \mathbf{\hat{x}}_d^{'}, \mathbf{\hat{x}}_j \right)
    = - \sum_j \frac{ \mathbf{\hat{n}}_d \cdot \mathbf{\hat{n}}_j}{\|\mathbf{\hat{p}}_d - \mathbf{\hat{p}}_j\|^3} \cdot \left(\mathbf{\hat{p}}_d - \mathbf{\hat{p}}_j\right) \label{eq:ppa_jacob_pos} \\
    & \frac{\partial}{\partial \mathbf{\hat{n}}_d} \sum_j \mathbf{ppa} \left( \mathbf{\hat{x}}_d^{'}, \mathbf{\hat{x}}_j \right)
    = \sum_j
    \frac{\mathbf{\hat{n}}_j - \left(\mathbf{\hat{n}}_d \cdot \mathbf{\hat{n}}_j\right) \cdot \mathbf{\hat{n}}_d}
    {\|\mathbf{\hat{p}}_d - \mathbf{\hat{p}}_j\|} \label{eq:ppa_jacob_ori}
\end{align}

We provide calculation details in Appendix.
By calculating the Jacobian vector, we obtain the gradient of PPA values with respect to the drone's poses and update our drone's pose by following its direction and moving by a step size $\Delta T$ in such a way that the constraints described in Eq.~\ref{eq:obj_func_ppa} are satisfied. 

\begin{equation}
    \begin{split}
        \mathbf{\hat{x}}_d^{'} & = \mathbf{\hat{x}}_d + \frac{\partial}{\partial \mathbf{\hat{x}}_d} \sum_j \mathbf{ppa} \left( \mathbf{\hat{x}}_d^{'}, \mathbf{\hat{x}}_j \right) \cdot \Delta T   \\
        \text{s.t. }    \quad &   \| \mathbf{\hat{p}}_d^{'} - \mathbf{\hat{p}}_d \| \leq T \text{,} \quad \|\mathbf{\hat{p}}_d^{'} - \mathbf{p}_a\| \geq R_{\text{safe}}
    \end{split}
\end{equation}

We provide the pseudo-code as below in Algo.~\ref{algo:local_vp_algorithm}. We use simple geometry primitives such as cuboids to model the actor during online planning. We use five faces (excluding the bottom) as the patches and build them according to the actor's 2D pose. We will show that this would produce similar reconstruction quality in Sec.~\ref{sec:exp_reconstruction} while reducing computational costs.

\begin{algorithm}
    \caption{Local View planning algorithm (\textbf{Local\_VP})}
    \label{algo:local_vp_algorithm}
    \begin{algorithmic}[1]
        \Require Drones' pose estimation $\mathbf{\hat{x}}_{d}$ and an actor's observation
        
        \While {planning}
        \State $\mathcal{X}_{est} \gets \{\mathbf{\hat{x}}_{j}\}$. Build actors' reconstruction estimation based on the observation.
        
        
        \State $\mathbf{v} \gets \frac{\partial}{\partial \mathbf{\hat{x}}_d} \sum_j \mathbf{ppa} \left( \mathbf{\hat{x}}_d^{'}, \mathbf{\hat{x}}_j \right)$. Calculate the jacobian vector of the PPA value.
        \State $\Delta t \gets \text{Constraints}(\mathbf{\hat{x}}_d, \mathbf{v}, R_{safe}, T)$. Calculate the step size while satisfying the constraints in Eq.~\ref{eq:obj_func_ppa}.
        \State $\mathbf{\hat{x}}_d^{'} \gets \mathbf{\hat{x}}_d + \Delta t \cdot \mathbf{v}$ \quad Calculate the target view point.
        \State $\mathbf{\hat{x}}_d \gets \mathbf{\hat{x}}_d^{'}$ \quad Execute and update camera's pose estimation.
        \EndWhile
        
        \State \Return $\mathcal{X}_{est}$
    \end{algorithmic}
\end{algorithm}

\section{SYSTEM DESIGN} \label{sec:system}
\begin{figure*}[th]
    \centering
    \includegraphics[width=1.0\textwidth]{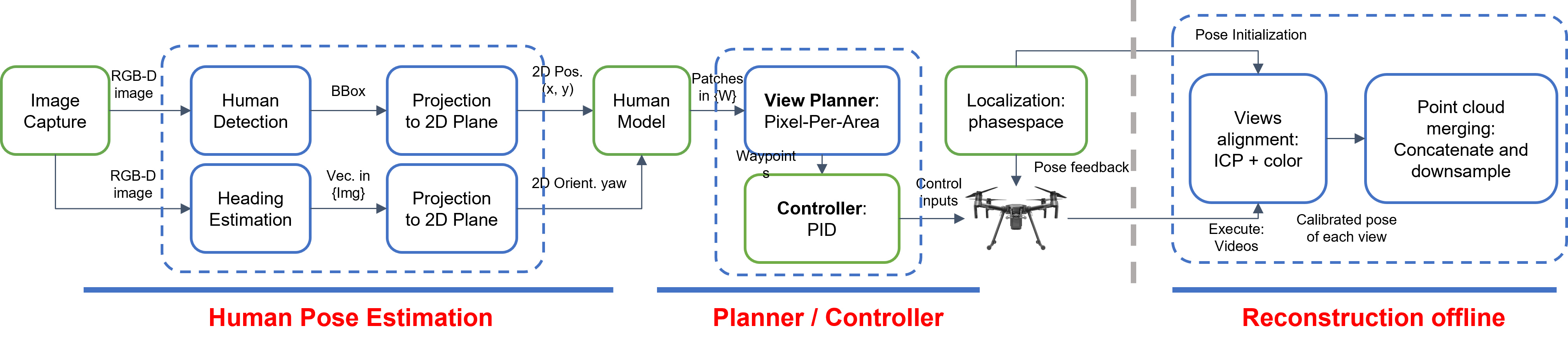}
    \caption{\textbf{System design.} Our system consists of three main modules. The online localization and heading direction estimation module, the online view planning module, and the offline 3D reconstruction module. Green boxes are executed on the onboard computing units. Blue ones are executed on the base station.}
    \label{fig:pipeline}
\end{figure*}

To validate our view planning in a real system, we build the drone system as shown in Fig.~\ref{fig:pipeline}. Besides our online view planning algorithm for high fidelity reconstruction in Sec.~\ref{sec:view_planning}, the system also includes the capabilities of i)~Online actor localization and heading direction estimation. ii)~Offline reconstruction with Iterative Closest Point (ICP) method.
We assume the actor is on the ground and localize it on its 2D pose in Sec.~\ref{sec:2D_pose_est}. Then we build human patches based on the 2D pose and plan views as described in Sec.~\ref{sec:view_planning}. We use RGB-D images from onboard cameras to produce high-fidelity reconstruction and calibrated views with the Iterative Closest Point method in Sec.~\ref{sec:reconstruction}.

\subsection{Online localization and heading direction estimation} \label{sec:2D_pose_est}

\begin{figure}[htb]
    \centering
    \includegraphics[width=0.75\columnwidth]{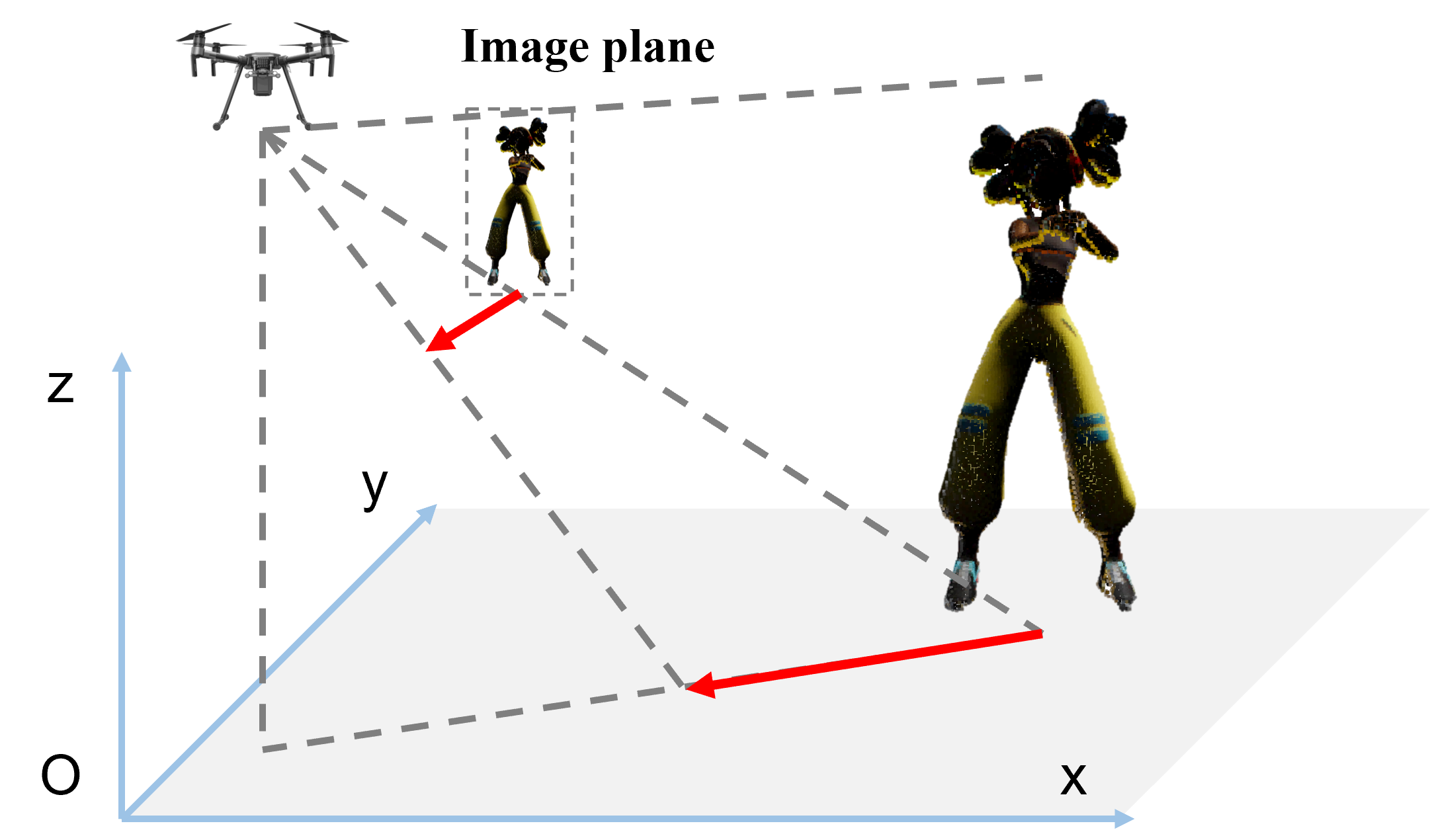}
    \caption{\textbf{2D position localization and heading direction estimation.} In order to localize the actor in real time, we use the middle point from the bounding box's bottom edge as its position representation in the image plane. We back-project from the image plane to the ground plane to estimate the position and the heading direction.} 
    \label{fig:projection}
\end{figure}

We follow the method proposed by Wenshan et al. \cite{wang_improved_2019} to obtain the actor's 2D pose. We show the actor localization and heading direction estimation (HDE) in Fig.~\ref{fig:projection}. Following the notation in Sec.~\ref{sec:formulation}, we denote actor's pose $\mathbf{x}_a$ in 2D ground plane as $(x_a, y_a, \theta_a)$, where $\theta_a$ denotes the actor's yaw angle, i.e. $\mathbf{p}_a = (x_a, y_a)$, $\mathbf{n}_a = \left(\cos{\theta_a}, \sin{\theta_a}\right)$. 

\subsubsection{Localization from bounding boxes} \label{sec:localization}

We extract the actor's bounding box in the image frame. We use the middle point of the bottom edge as the actor's position projection in the image plane, and we back-project it to the ground plane to estimate the actor's 2D position.

\subsubsection{Heading Direction Estimation from 2D skeleton pose}

Given an actor's image, we extract the human skeleton pose with AlphaPose \cite{fang_rmpe_2017, xiu_pose_2018} as a $17 \times 2$ vector, and feed them into a Multiple Layer Perception (MLP). The output of the network is a unit vector in the image frame $\left(\cos{\theta_{img}}, \sin{\theta_{img}}\right)$. Similarly, we back-project it to the ground plane to obtain its yaw estimation.
 

To train the MLP network, we use five videos (around 750 training frames) recorded from a static camera as our training data. We require our actor to move along its heading direction. We use the moving direction of the actor's bounding box as the ground truth. We define the loss function as $L = 1 - \cos{\Delta \theta}$, where $\Delta \theta$ is the angle between the prediction and the ground truth $\theta_{img}$.

We use a Kalman Filter \cite{welch_introduction_1995} to update the 2D poses from multiple estimations during the flight. We assume a uniform distribution of our actor's movement and propagate the actor's position with a constant speed.

\subsection{Offline reconstruction with Iterative Closet Point (ICP)} \label{sec:reconstruction}

While executing planned views, we use the onboard camera to record RGB-D images and obtain high-fidelity reconstruction offline. While the actor is dynamic, we use a few consecutive views (2-5 frames) to reconstruct them by assuming the actor's motion is small in less than half a second. We use Colored Iterative Closest Point (ICP) \cite{chen_object_1992, rusinkiewicz_efficient_2001, park_colored_2017}  method to align consecutive views, merge them by concatenating point clouds and downsample them by each result.

\subsection{System information.}
We use DJI M100 as our working drone, with Jetson TX1 as our computing unit and Realsense D435 as our onboard camera. Due to the computational constraints, we record RGB-D images at 15Hz. We fix markers on the drone and use the Phasespace motion capture system \cite{phasespace_inc_optical_nodate} to localize our drone during the flight. To provide additional computation resources, we use a base computer communicating online with Jetson to extract human skeleton poses and plan views through ROS system \cite{stanford_artificial_intelligence_laboratory_et_al_robotic_2018}. We guarantee our drone and the actor's safety by checking planned views with a safety module.

\section{EXPERIMENTS} \label{sec:experiments}
We study the effectiveness of our system through the following questions:

\begin{enumerate}
    \item How does the PPA function perform as a proxy for the reconstruction quality?
    \item How is the reconstruction quality improved by optimizing the PPA metric?
    \item How accurate is our actor localization and heading direction estimation module?
\end{enumerate}

\begin{figure}[th]
    \centering
    \includegraphics[width=1.0\columnwidth]{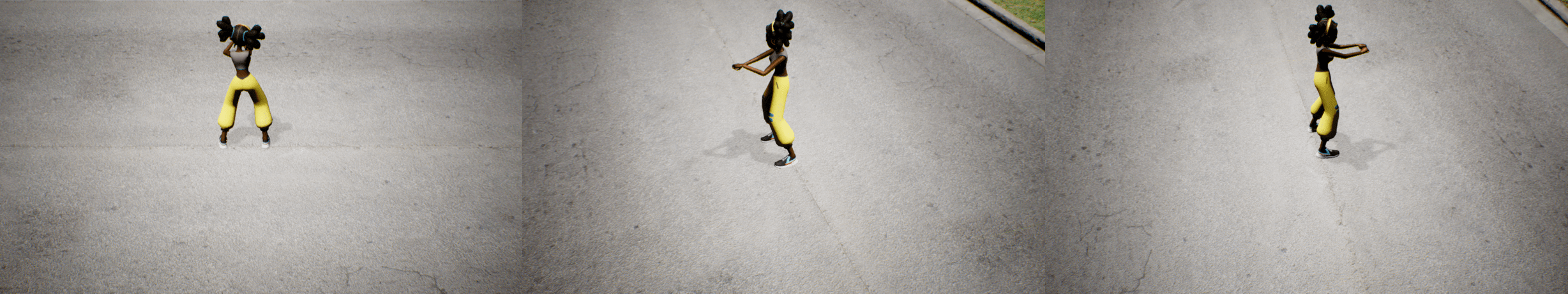}
    \caption{\textbf{Airsim simulation in Unreal Engine.} We use Airsim simulation to obtain the ground truth of the human actor surface and compare it against our reconstruction results. We use the Computer Vision mode of the Airsim, where the images of the actor are taken in the first-person view of the camera.}
    \label{fig:exp_airsim}
\end{figure}

To answer these questions, we conduct experiments both in real-world setups and simulations. We use Microsoft Airsim~\cite{shah_airsim_2017} and Unreal Engine~\cite{epic_games_unreal_2019} simulation environment. We use human animations from Mixamo~\cite{adobe_systems_inc_mixamo_nodate} to evaluate each method's reconstructions (Fig.~\ref{fig:exp_airsim}).

In the simulation, we set the camera Field-of-View as $90^\circ$. RGB-D images are in $1920\times1080$ resolution. Depth ground truth and segmentation are generated from Airsim API. We share all our code for the simulation in the \href{https://github.com/Qingyuan-Jiang/view_planning_3dhuman}{link}.

\subsection{PPA as the reconstruction quality proxy} \label{sec:exp_ppa}

In this part, we show the correlation between PPA values and reconstruction quality and the validity using geometry primitives to represent a human actor.

\subsubsection{Experiment setup in the simulation.}
We conduct experiments in Airsim for the actor's surface ground truth, using the mesh as the representation. We use triangles and their surface normals as our patches $\mathbf{x}_j$. We sample viewpoints uniformly in $r$, $\theta$, and $\phi$ in spherical coordinates in the actor frame. We calculate the PPA values for each view by averaging among all visible patches. We also calculate the corresponding PPA values if we model the actor as a cuboid, i.e., five faces as patches. We compare them with the reconstruction quality metric defined below.

\textbf{i) Coverage of triangles}. We project each pixel from the sampled view into 3D space with depth information. We define a point from a point cloud inside the triangular prism if it is located within a fixed height. We define a triangle as visible if at least one point is in the prism. In our experiments, we fix the height of the prism as $1cm$. Since only parts of the actor can be seen from a single view, we use the percentage of visible triangles to compute the coverage rate. \textbf{ii) Average Pixels-Per-Triangle}. We also use the average number of pixels on a mesh triangle as our evaluation metric.

\subsubsection{Results}
\begin{figure}[th]
    \centering
    \includegraphics[width=0.9\columnwidth]{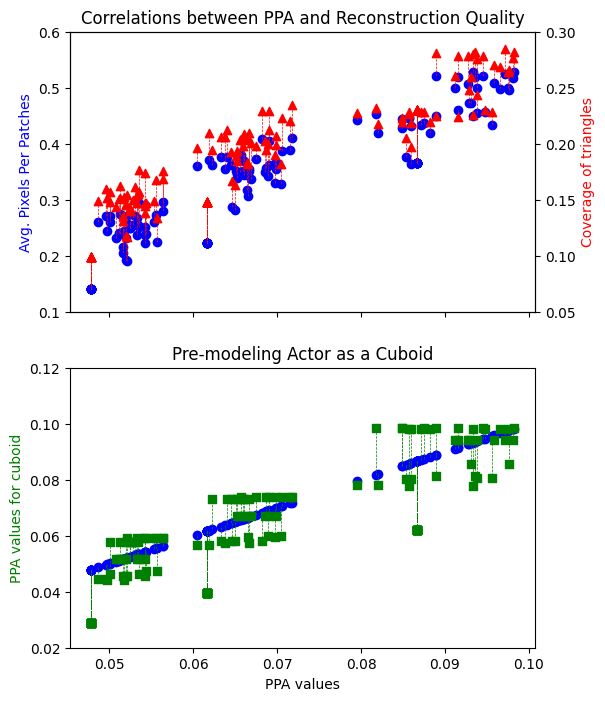}
    \caption{\textbf{Correlation between PPA values and the reconstruction quality.} We show the correlation between PPA values and reconstruction quality and validity to use geometry primitives as the actor model. X-axis for both figures: PPA values on each human mesh triangle. Y-axis (top): metrics for reconstruction quality, average pixels per patch in blue dots, coverage ratio in red triangle. Y-axis (bottom): PPA values based on the cuboid model. Blue dots are the PPA values on each mesh triangle and hence on $y=x$ diagonal. The corresponding PPA values on the cuboid faces are connected to the green dots.}
    \label{fig:exp_correlation}
\end{figure}

Quantitative results are shown in Fig~\ref{fig:exp_correlation}. Each dot is a sampled view. We visualize the PPA values from mesh triangles as the X-axis. In the top figure of Fig.~\ref{fig:exp_correlation}, we visualize two metrics as the Y-axis. We show the monotonic increase between PPA values and two metrics, which implies the correlation between PPA values and the reconstruction quality.

The bottom figure shows PPA values based on mesh triangles on the Y-axis in blue and PPA values based on cuboid representation in green. 
The results indicate that geometry primitive such as cuboid would provide a similar effect as the ground truth regarding the reconstruction quality. This is intuitive because many patches are correlated within part of the body area, for example, for patches on the chest or back. The results show that it is reasonable to simplify them as one for view planning purposes.

\subsection{PPA based view planning strategy} \label{sec:exp_reconstruction}

In this part, we compare the reconstruction results generated from various baselines and show that by optimizing PPA, we obtain better reconstruction quality compared to the other view planning baselines.

\subsubsection{Metric} We use the coverage ratio as our evaluation metric. In addition, we use the Chamfer Distance (CD) ~\cite{fan_point_2017} as our second evaluation metric.
$\text{CD}(X, Y) = \frac{1}{|X|} \sum_{\mathbf{x}_i \in X} \min_{\mathbf{y}_j \in Y} ||\mathbf{x}_i - \mathbf{y}_j||_2$.
Chamfer distance in the forward direction computes the accuracy of the reconstruction, whereas, in the backward direction, it models the coverage of the ground truth point cloud by the reconstruction. We report the mean of the two directions as the total reconstruction error. All numbers are reported in millimeters.
The Chamfer Distance can be relatively small compared to the real robot scenario because we use the ground truth of the actor surface from the simulation.

\subsubsection{Baselines} Similar to Sec.~\ref{sec:exp_ppa}, we sample views around the actor as initial view poses. 
We compare our method with mesh-based PPA planning, where we optimize our PPA values based on partial point cloud observation from the ground truth. We also compare with the greedy method, where the drone moves to the closest position while maintaining a safe distance and the same viewing direction. Our third baseline is set to enumerate the viewing quality metrics in the local area. We use it as our upper bound, which can not be reached during actual flights since we can not obtain the mesh ground truth of the actor surface. We set the safe radius of the human actor as $R_{safe} = 8$m and step size as $T = 1.0$m in the experiments.

\begin{table*}[ht]
    \begin{center}
        \caption{Reconstruction results from view planning algorithms.}
        \begin{tabular}{|l|c|c|c|c|c|c|}
            \hline
            \textbf{Modules}        & No Plan   & Greedy  & PPA Cuboid & PPA Mesh & Enum. Coverage & Enum. CD. \\
            \hline
            Coverage [\%]$\uparrow$ & 12.6  & 12.8  & 12.9  & \textbf{13.7} & 14.2  & 13.8\\
            \hline
            CD [mm] $\downarrow$    & 45.36 & 44.72 & 44.65 & \textbf{44.24}& 43.65 & 43.10\\
            \hline
            Coverage (noise) [\%]$\uparrow$ & 12.6  & 12.9  & 13.0  & \textbf{13.6}  & 14.1  & 13.6\\
            \hline
            CD (noise) [mm] $\downarrow$    &45.36  & 45.25 & 45.08 & \textbf{44.46} & 43.80 & 43.13\\
            \hline
        \end{tabular}
    \label{table:exp_reconstruction_error}
    \end{center}
\end{table*}

\begin{figure}[th]
    \centering
    \includegraphics[width=1.0\columnwidth]{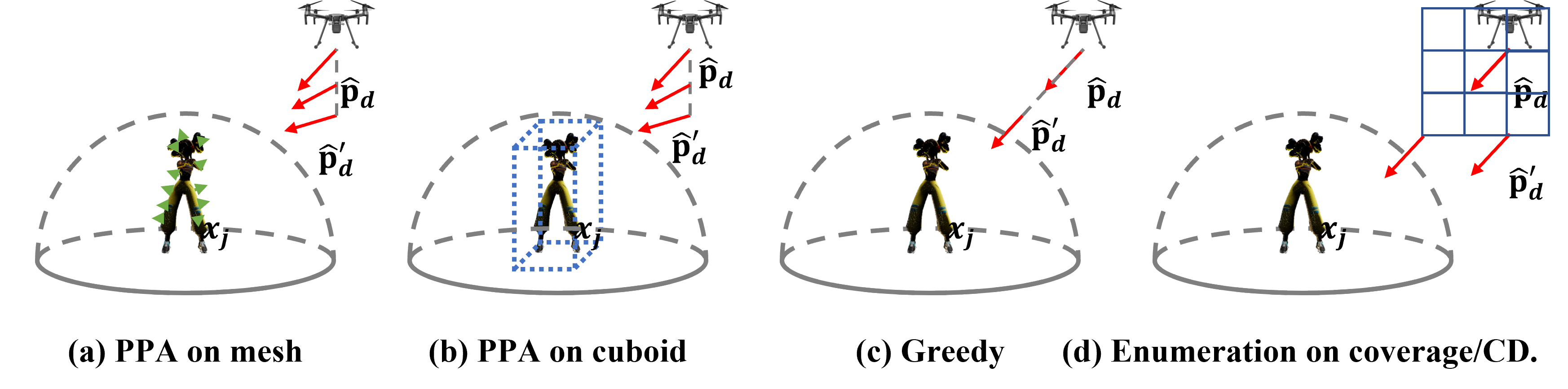}
    \caption{\textbf{Methods and baselines} We compare our planning strategy with other baselines on the reconstruction results. From left to right, we show the view planning methods, including PPA based on the mesh, PPA based on a cuboid, greedy method, and enumeration on viewing quality.}
    \label{fig:exp_baselines}
\end{figure}

\subsubsection{Analysis} From the results in Table~\ref{table:exp_reconstruction_error}, we show that optimizing our PPA values can adjust viewpoints by covering more details and lower reconstruction error. Meanwhile, we also show that optimizing PPA values based on whole mesh triangles would generate better reconstruction results than the cuboid-based actor model. 
It would be ideal for adjusting viewpoints based on high-fidelity reading, such as point cloud from RGB-D camera, during the flying time, with increased computational cost. Modeling the actor as a cuboid may generate viewpoints with similar reconstruction quality but lower computational consumption.

\subsubsection{Sensibility test} Besides, we test the sensitivity of the view planning algorithm when the error of human 2D pose is included. Gaussian noise ($\mu=0$m, $\sigma=0.5$m) is added to the position and yaw angle of the actor pose regarding the greedy method and PPA cuboid method. From Table~\ref{table:exp_reconstruction_error}, we show our view planning method is robust with respect to the localization noise.

\subsubsection{Qualitative results from real world}

\begin{figure}[ht]
    \centering
    \includegraphics[width=1.0\columnwidth]{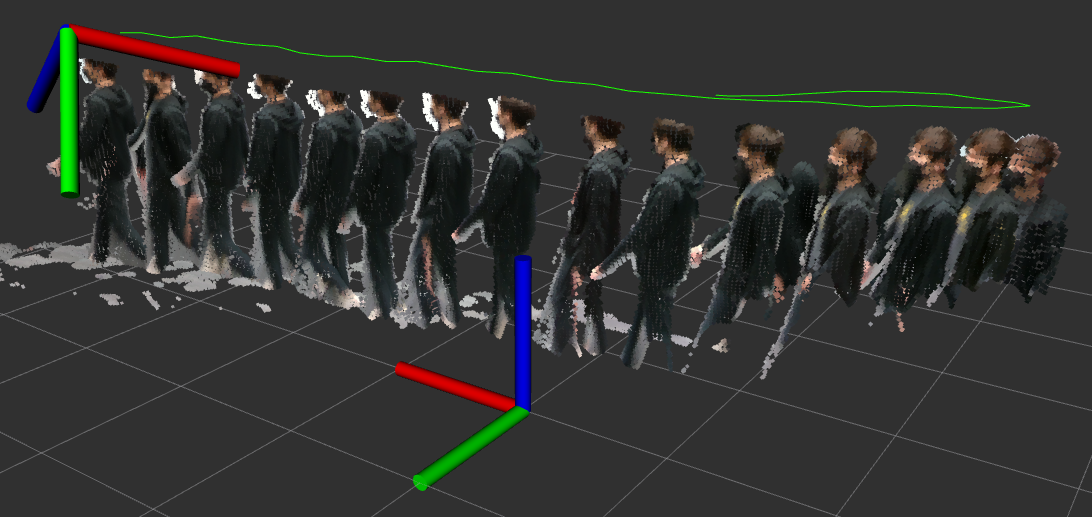}
    \caption{\textbf{Reconstruction results from real drone setup.} We show our reconstructed human actor from the real drone in ROS Rviz. Green lines are the calibrated camera path. Results are shown as the point cloud in the middle. The coordinate frame in the middle (resp. top left) corresponds to the world (resp. camera) frame.
    \label{fig:reconstruction}}
\end{figure}

We also show qualitative results from the real-world experiments in  Fig.~\ref{fig:reconstruction}. We visualize the dynamics of the actor walking in the 3D world. We show the world coordinate as the frame in the middle and the camera frame at the last view in the left-top corner. We show the camera path in a green line and reconstruct the walking actor from right to left. A full video of the planned views and reconstructed actor is included in the supplementary video.

\subsection{Overall System Performance} \label{sec:exp_hde}

We conduct quantitative experiments to measure the accuracy of our localization and heading direction estimation system in Sec.~\ref{sec:localization}.

\subsubsection{Experiment setup}
To obtain the ground truth of the actor's 2D pose, we put markers of Phasespace on the actor. 
We compare it with our estimation of position and yaw angle. We use Mean Squared Error (MSE) as our evaluation metric for the position estimation and average absolute degree difference as our metric for the yaw estimation.


\subsubsection{Results}
Results show that we can localize the actor within $0.63$m in position estimation and $58.28^{\circ}$ in HDE.
Meanwhile, each module can run in $60.0$ms and $10.95$ms separately, which can serve in real time.
The error comes from 1) Bounding box inaccuracy. We currently use the middle point of the bounding box's bottom edge to represent the actor's position, which may not be accurate in some edge cases. 2) Parameters influenced by the K.F. prediction. We assume our actor moves within the $1$m/s range, which may enlarge our covariance in the K.F. propagation.


\section{CONCLUSION} \label{sec:conclusion}
In this paper, we presented a method to plan views of a UAV system so as to acquire images of a dynamic actor, which are then used for obtaining high-fidelity reconstructions of the actor. We formulated a view planning problem and solved it by defining the Pixels-Per-Area (PPA) function as our reconstruction quality proxy. We used a simple geometric representation for the actor model to avoid costly object representations. We integrated our drone system with online actor localization and heading direction estimation module. Putting these components together, we produced pixel-level 3D reconstructions by tracking the actor's 2D pose while executing planned views online and merging the offline RGB-D camera readings into the point cloud. We conducted experiments in both simulation and a real-world setup. We examined and established the effectiveness of PPA as a proxy by measuring the correlation between PPA and reconstruction quality. We showed that with the proposed algorithm, the drone could capture the 3D reconstruction results in better quality. We also demonstrated our reconstruction results of the dynamic actor from the real world.

In this work, we performed robot experiments in our indoor drone lab to quantitatively evaluate the system. In the future, we would like to extend our results to  outdoor environments where localization errors might hinder the system's performance and therefore incorporate them into the view planning strategy.

\section{DISCUSSION} \label{sec:discussion}
In this work, we focus on the view planning algorithm for a single dynamic actor. The problem becomes more difficult when there are multiple actors, which brings more patches to be seen while occlusions are in between. To generalize, we may formulate the problem as a Traveling Salesperson Problem with Neighborhoods (TSP-N), where we define a viewpoint $\mathbf{r}_{ij}$ for each patch-$j$ and define each neighborhood by setting a threshold to the PPA metric, as in Eq.~\ref{eq:obj_func_tspn}.

\begin{equation} \label{eq:obj_func_tspn}
    \begin{split}
    \min_{\mathbf{r}_{ij}, \left(j_1 \ldots j_k \ldots j_m\right)} \quad L & = \sum_{k=0}^{m-1} dist \left( \mathbf{r}_{i{j_k}}, \mathbf{r}_{i{j_{k+1}}} \right) \\
    \text{s.t. }    \quad &   \mathbf{ppa} \left( \mathbf{r}_{ij}, \mathbf{p}_j, \mathbf{n}_j \right) \geq C \\
    &   d(\mathbf{r}_{ij}, \mathbf{p}_j) \geq R_{\text{safe}}   
    \end{split}
\end{equation}

We will not go into detail on this scenario since we do not experiment in the real-world drone system due to the spatial limitation of our drone room. We will plan this experiment as our future work in the outdoor environment.

\section*{APPENDIX}
In the appendix, we will first discuss the geometric meaning of the PPA metric and then further explain the math behind it.

\subsection{Geometric Meaning of PPA} \label{sec:appendix_geo_meaning}

\begin{figure}[ht]
    \centering
    \includegraphics[width=0.6\columnwidth]{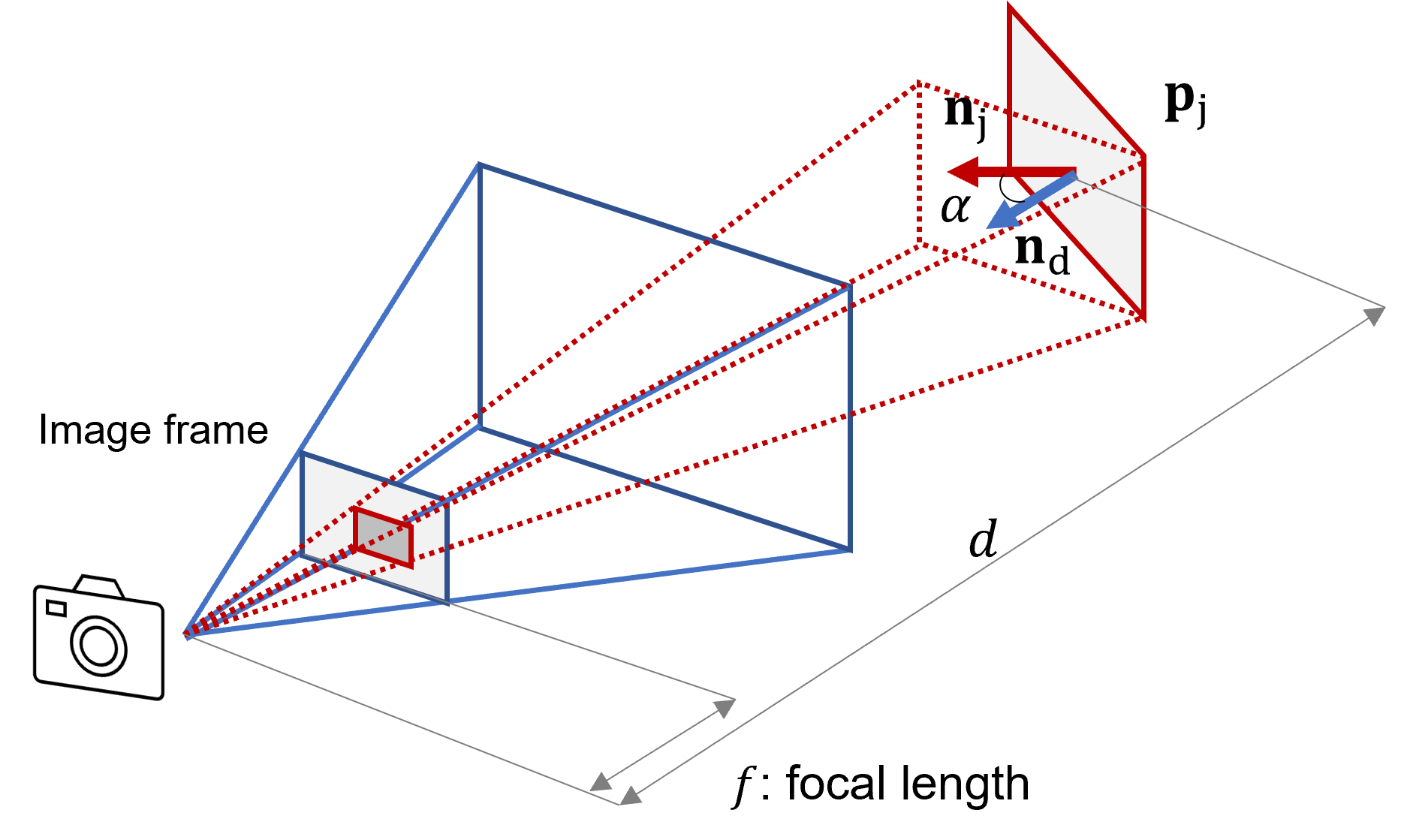}
    \caption{\textbf{Geometric meaning of PPA values}. A patch (colored in \textcolor{red}{red}) is projected into the image plane (colored in \textcolor{blue}{blue}. $f$ stands for the focal length of a camera.}
    \label{fig:ppa_geometric_meaning}
\end{figure}

As defined in Sec.~III-A, we use the projection length in the image plane of the 3D segment, or in other words, the ratio between the area of a 3D patch and its projection in the image plane, as illustrated in the Fig.~\ref{fig:ppa_geometric_meaning}.

\begin{equation}
    \begin{split}
        \mathbf{ppa} \left( \mathbf{x}_d, \mathbf{x}_j \right) 
        &= \frac{A_{img}}{A_j} = \frac{A_{img}}{A_{norm}} \frac{A_{norm}}{A_{j}} \\
        &= \frac{f}{d(\mathbf{p}_d, \mathbf{p}_j)} \cos(\alpha(\mathbf{n}_d, \mathbf{n}_j)) \\
        &= \left(\frac{\cos(\alpha(\mathbf{n}_d, \mathbf{n}_j))}{d(\mathbf{p}_d, \mathbf{p}_j)} \right) f
    \end{split}
\end{equation}

Since the focal length of a camera is a constant, we define the PPA function as Eq.~2. 

\subsection{Jacobian of PPA} \label{sec:appendix_jacobian}

In Sec.~IV, we propose the view planning algorithm based on optimizing the PPA functions. The underlying assumption is that the camera's orientation is de-coupled with its position. As stated in the definition of PPA Eq.2 and the corresponding jacobian Eq.~5. The entire equation could be written as follows.

\begin{equation}    \label{eq:ppa_jacob_pos_full}
    \begin{split}
        &\frac{\partial}{\partial \mathbf{\hat{p}}_d^{'}} \sum_j \mathbf{ppa} \left( \mathbf{\hat{x}}_d^{'}, \mathbf{\hat{x}}_j \right) \\
        &= \frac{\partial}{\partial \mathbf{\hat{p}}_d^{'}} \sum_j \frac{\cos(\alpha(\mathbf{\hat{n}}_d^{'}, \mathbf{\hat{n}}_j))}{d(\mathbf{\hat{p}}_d^{'}, \mathbf{\hat{p}}_j)} \\
        &= \frac{\partial}{\partial \mathbf{\hat{p}}_d^{'}} \sum_j \left[
        \frac{\mathbf{\hat{n}}_d^{'} \cdot \mathbf{\hat{n}}_j}{\| \mathbf{\hat{n}}_d^{'} \| \cdot \|\mathbf{\hat{n}}_j\|}
        \frac{1}{\| \mathbf{\hat{p}}_d^{'} - \mathbf{\hat{p}}_j \|} \right]  \\
        &= \sum_j (\mathbf{\hat{n}}_d^{'} \cdot \mathbf{\hat{n}}_j) 
        \frac{\partial}{\partial \mathbf{\hat{p}}_d^{'}}
        \frac{1}{\| \mathbf{\hat{p}}_d^{'} - \mathbf{\hat{p}}_j \|}  \\
        &= \sum_j (\mathbf{\hat{n}}_d^{'} \cdot \mathbf{\hat{n}}_j)
        \frac{-1}{\| \mathbf{\hat{p}}_d^{'} - \mathbf{\hat{p}}_j \|^2}
        \frac{\partial}{\partial \mathbf{\hat{p}}_d^{'}}
        \| \mathbf{\hat{p}}_d^{'} - \mathbf{\hat{p}}_j \| \\
        &= \sum_j (\mathbf{\hat{n}}_d^{'} \cdot \mathbf{\hat{n}}_j)
        \frac{-1}{\| \mathbf{\hat{p}}_d^{'} - \mathbf{\hat{p}}_j \|^2}
        \frac{\mathbf{\hat{p}}_d^{'} - \mathbf{\hat{p}}_j}{\| \mathbf{\hat{p}}_d^{'} - \mathbf{\hat{p}}_j \|} \\
        &= - \sum_j \frac{ \mathbf{\hat{n}}_d^{'} \cdot \mathbf{\hat{n}}_j}{\|\mathbf{\hat{p}}_d^{'} - \mathbf{\hat{p}}_j\|^3} \cdot \left(\mathbf{\hat{p}}_d^{'} - \mathbf{\hat{p}}_j\right) \\
    \end{split}
\end{equation}

Similarly, given $\|\mathbf{\hat{n}}_d\| = \|\mathbf{\hat{n}}_j\| = 1$, we have

\begin{equation}    \label{eq:ppa_jacob_ori_full}
    \begin{split}
        &\frac{\partial}{\partial \mathbf{\hat{n}}_d^{'}} \sum_j \mathbf{ppa} \left( \mathbf{\hat{x}}_d^{'}, \mathbf{\hat{x}}_j \right) \\
        &= \frac{\partial}{\partial \mathbf{\hat{n}}_d^{'}} \sum_j \left(
        \frac{\mathbf{\hat{n}}_d^{'} \cdot \mathbf{\hat{n}}_j}{\| \mathbf{\hat{n}}_d^{'} \| \cdot \|\mathbf{\hat{n}}_j\|}
        \frac{1}{\| \mathbf{\hat{p}}_d^{'} - \mathbf{\hat{p}}_j \|} \right)  \\
        &= \sum_j \frac{1}{\| \mathbf{\hat{p}}_d^{'} - \mathbf{\hat{p}}_j \|} 
        \frac{\partial}{\partial \mathbf{\hat{n}}_d^{'}} 
        \left(
        \frac{\mathbf{\hat{n}}_d^{'} \cdot \mathbf{\hat{n}}_j}{\| \mathbf{\hat{n}}_d^{'} \| \cdot \|\mathbf{\hat{n}}_j\|}
        \right) \\
        &= \sum_j \frac{1}{\| \mathbf{\hat{p}}_d^{'} - \mathbf{\hat{p}}_j \|} 
        \frac{\mathbf{\hat{n}}_j - (\mathbf{\hat{n}}_d^{'} \cdot \mathbf{\hat{n}}_j) \frac{\mathbf{\hat{n}}_d^{'}}{\|\mathbf{\hat{n}}_d^{'}\|}\cdot \|\mathbf{\hat{n}}_j\|}{\| \mathbf{\hat{n}}_d^{'} \|^2 \cdot \|\mathbf{\hat{n}}_j\|^2}  \\
        &= \sum_j \frac{\mathbf{\hat{n}}_j - \left(\mathbf{\hat{n}}_d \cdot \mathbf{\hat{n}}_j\right) \cdot \mathbf{\hat{n}}_d}{\|\mathbf{\hat{p}}_d - \mathbf{\hat{p}}_j\|}
    \end{split}
\end{equation}

However, for a single-actor case, since the drone camera is relatively far from the actor patch, we can simplify the problem by constraining the camera orientation targeting the actor, i.e., 

\begin{equation} \label{eq:ppa_additional_cons}
    \mathbf{n_d} = \frac{\mathbf{p_a} - \mathbf{p_d}}{\|\mathbf{p_a} - \mathbf{p_d}\|}
\end{equation}

With such constraints, we could rewrite our PPA functions and corresponding Jacobian matrix as below.

\begin{equation} \label{eq:ppa_define_append}
    \begin{split}
        \mathbf{ppa} \left( \mathbf{\hat{x}}_d^{'}, \mathbf{\hat{x}}_j \right)
        & = \frac{\cos(\alpha(\mathbf{\hat{n}}_d^{'}, \mathbf{\hat{n}}_j))}{d(\mathbf{\hat{p}}_d^{'}, \mathbf{\hat{p}}_j)}  \\
        & = \frac{\mathbf{\hat{n}}_d^{'} \cdot \mathbf{\hat{n}}_j}{\| \mathbf{\hat{n}}_d^{'} \| \cdot \|\mathbf{\hat{n}}_j\|} \frac{1}{ \| \mathbf{\hat{p}}_d^{'} - \mathbf{\hat{p}}_j \|} \\
        & = \frac{(\mathbf{\hat{p}}_d^{'} - \mathbf{\hat{p}}_j) \cdot \mathbf{\hat{n}}_j}{\| \mathbf{\hat{p}}_d^{'} - \mathbf{\hat{p}}_j \| \cdot \| \mathbf{\hat{n}}_d^{'} \| \cdot \|\mathbf{\hat{n}}_j\|} \frac{1}{ \| \mathbf{\hat{p}}_d^{'} - \mathbf{\hat{p}}_j \|} \\
        & = \frac{(\mathbf{\hat{p}}_d^{'} - \mathbf{\hat{p}}_j) \cdot \mathbf{\hat{n}}_j}{ \| \mathbf{\hat{p}}_d^{'} - \mathbf{\hat{p}}_j \|^2} \\
    \end{split}
\end{equation}

\begin{equation}    \label{eq:ppa_jacob_pos_cons_full}
    \begin{split}
        &\frac{\partial}{\partial \mathbf{\hat{p}}_d^{'}} \sum_j \mathbf{ppa} \left( \mathbf{\hat{x}}_d^{'}, \mathbf{\hat{x}}_j \right) \\
        & = \sum_j \frac{\partial}{\partial \mathbf{\hat{p}}_d^{'}} \frac{(\mathbf{\hat{p}}_d^{'} - \mathbf{\hat{p}}_j) \cdot \mathbf{\hat{n}}_j}{ \| \mathbf{\hat{p}}_d^{'} - \mathbf{\hat{p}}_j \|^2} \\
        & = \sum_j \frac{\mathbf{\hat{n}}_j \cdot (\mathbf{\hat{p}}_d^{'} - \mathbf{\hat{p}}_j)^2 - ((\mathbf{\hat{p}}_d^{'} - \mathbf{\hat{p}}_j) \cdot \mathbf{\hat{n}}_j) \cdot 2 (\mathbf{\hat{p}}_d^{'} - \mathbf{\hat{p}}_j) }
        { \| \mathbf{\hat{p}}_d^{'} - \mathbf{\hat{p}}_j \|^4} \\
        & = \sum_j \frac{\mathbf{\hat{n}}_j \cdot \| \mathbf{\hat{p}}_d^{'} - \mathbf{\hat{p}}_j \| - \left(\frac{\mathbf{\hat{p}}_d^{'} - \mathbf{\hat{p}}_j}{ \| \mathbf{\hat{p}}_d^{'} - \mathbf{\hat{p}}_j \|} \cdot \mathbf{\hat{n}}_j \right) \cdot 2 (\mathbf{\hat{p}}_d^{'} - \mathbf{\hat{p}}_j)}
        { \| \mathbf{\hat{p}}_d^{'} - \mathbf{\hat{p}}_j \|^3} \\
        & = \sum_j \frac{\mathbf{\hat{n}}_j \cdot d - 2\cos(\alpha) (\mathbf{\hat{p}}_d^{'} - \mathbf{\hat{p}}_j)}
        { \| \mathbf{\hat{p}}_d^{'} - \mathbf{\hat{p}}_j \|^3} \\
        & = \sum_j \frac{\mathbf{\hat{n}}_j \cdot \frac{d}{2\cos(\alpha)} - (\mathbf{\hat{p}}_d^{'} - \mathbf{\hat{p}}_j)}
        { \| \mathbf{\hat{p}}_d^{'} - \mathbf{\hat{p}}_j \|^3} \\
    \end{split}
\end{equation}

\begin{figure}[th]
    \centering
    \includegraphics[width=0.8\columnwidth]{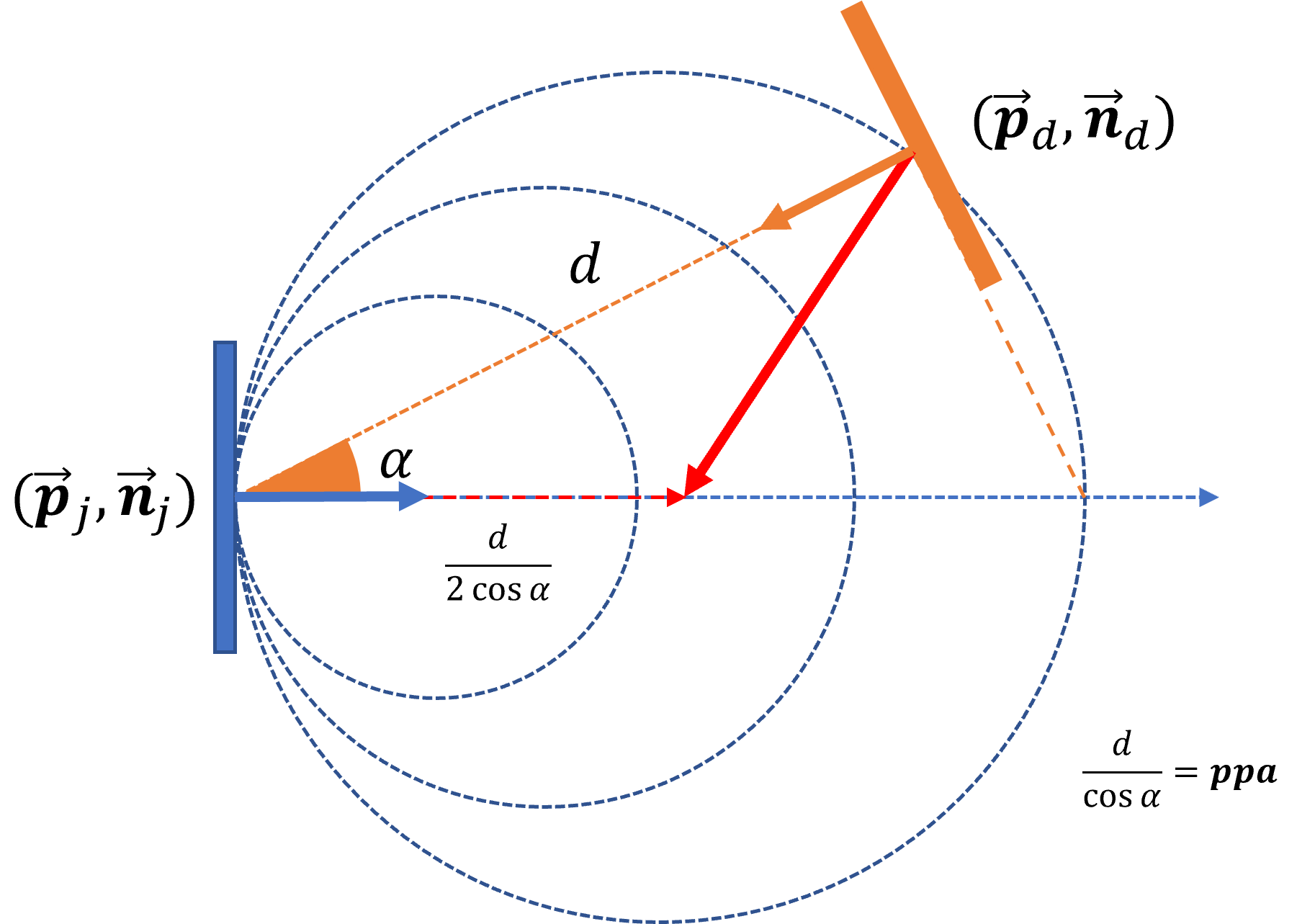}
    \caption{\textbf{Contour map of PPA values in 2D}. One actor patch is colored by \textcolor{blue}{blue}, and the drone pose is colored by \textcolor{orange}{orange}. PPA values' contour map is plotted by the \textcolor{blue}{blue dash}. The optimal direction to maximize the PPA value is drawn in \textcolor{red}{red}.}
    \label{fig:ppa_countour}
\end{figure}

The result is consistent with the geometric meaning. In Fig.~\ref{fig:ppa_countour}, we show the contour map of the PPA values with a single patch in a 2D situation. The contour of a certain PPA value $\mathbf{ppa} = t$ is a circle considering the equation $d = \cos(\alpha) t$. As drawn in Fig.~\ref{fig:ppa_countour}, the drone is located on the outside circle. The gradient of PPA values is the tangent direction which is pointing at the circle's center, as drawn in Fig.~\ref{fig:ppa_countour}, and can be described as $\mathbf{\hat{n}}_j \cdot \frac{d}{2\cos(\alpha)} - (\mathbf{\hat{p}}_d^{'} - \mathbf{\hat{p}}_j)$.

\section*{ACKNOWLEDGMENT}

The authors would like to thank Ruotong Wang for her emotional support. The author would also like to thank Selim Engin, Jun-Jee Chao, Pratik Mukherjee, Ishan Shetty, Ritik Mishra, Pranav Vijay, and all members of the Robotic Sensor Network Lab for their support. This work is supported by the NSF NRI Grant \#2022894.

\bibliographystyle{IEEEtran}
\bibliography{root.bib}

\end{document}